# Real-time Scene Text Detection Based on Global Level and Word Level Features

Fuqiang Zhao[1], Jionghua Yu[1], Enjun Xing[1,2], Wenming Song[1,2], and Xue Xu[1,2]

[1] Tianjin DM AILab, Tianjin , China
fqzhao@126.com
[2] Tianjin University of Finance and Economics, Tianjin , China
xingenjun@tjufe.edu.cn,xuxue2017@163.com

**Abstract.** It is an extremely challenging task to detect arbitrary shape text in natural scenes on high accuracy and efficiency. In this paper, we propose a scene text detection framework, namely GWNet, which mainly includes two modules: Global module and RCNN module. Specifically, Global module improves the adaptive performance of the DB (Differentiable Binarization) module by adding k submodule and shift submodule. Two submodules enhance the adaptability of amplifying factor k, accelerate the convergence of models and help to produce more accurate detection results. RCNN module fuses global-level and word-level features. The word-level label is generated by obtaining the minimum axis-aligned rectangle boxes of the shrunk polygon. In the inference period, GWNet only uses global-level features to output simple polygon detections. Experiments on four benchmark datasets, including the MSRA-TD500, Total-Text, ICDAR2015 and CTW-1500, demonstrate that our GWNet outperforms the state-of-the-art detectors. Specifically, with a backbone of ResNet-50, we achieve an F-measure of 88.6% on MSRA-TD500, 87.9% on Total-Text, 89.2% on ICDAR2015 and 87.5% on CTW-1500.

## 1 Introduction

Recently, arbitrary shape text detection natural scenes has attracted extensive attention in the computer vision field. The text detection algorithm aims to effectively locate the text regions in the natural image, so as to improve the recognition accuracy of the subsequent recognition system. The challenges of scene text reading mainly lie in varying orientations, scales and shapes of scene text. A recent popular trend is to perform scene text detection by segmentation-based methods because of its excellent description of various shapes, which benefitting from pixel-level segmentation.
However, most scene text detection methods do not focus on both the high accuracy and the inference speed. [29] proposed an Efficient and Accurate Scene Text Detector (EAST) to speed up the text detection, but the accuracy is relatively low and this method can not deal with irregular scene text. [30] propose one novel Fourier Contour Embedding (FCE) method to represent arbitrary shaped text contours as compact signatures. [3] propose to progressively evolve the initial text proposal to arbitrarily shaped text contours in a top-down manner. [11] proposed a Conditional Spatial Expansion (CSE) mechanism to improve the performance of curve text detection, which regards text detection as a region expansion process. [25] fused three levels of feature representations for text detection, which increases the detecting accuracy

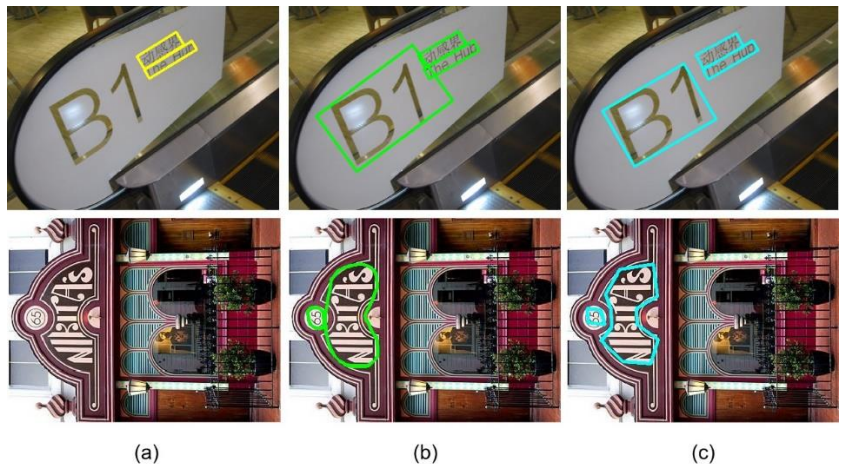
**Fig. 1.** Visualization of text detection from (a) DBNet, (b) GWNet and (c) ground truth.

and effectively detecting texts with arbitrary shapes by utilizing richer fused features. However, their inference speed is unable to meet real-time detection.

According to the shortcomings of existing text detection methods, an improved model called GWNet is proposed in this paper. It mainly includes Global module and RCNN module which fuse the text features from global level path(anchor-free) and word level path(anchor-based), respectively. Experimental results indicate that the proposed method outperforms DBNet on accuracy with the real time inference speed.

Figure 1 is a partial visualization of DBNet and GWNet on the testing datasets. It can be seen that GWNet has a certain improvement in detection performance, especially for some small or large scale text detection.

The contribution of our work is four-fold: (1) We propose a new text detector called GWNet. During the training period, we extract and fuse global-level and word-level features to enhance the detection accuracy. In the inference period, we perceive texts from global-level to improve the detecting efficiency; (2) We added k submodule and shift submodule to further optimize the adaptive performance of binarization process. GWNet can provide a highly robust binarization map; (3) We will introduce the word-level label for RCNN module, which is generated by obtaining the minimum axis-aligned rectangle boxes of the shrunk polygon; (4) GWNet achieves state-of-the-art performance on three benchmarks for ensuring both detection speed and accuracy.

## 2   Related Work

Recently, object detection/segmentation approaches based on deep learning, such as Faster R-CNN [4] and FCN [12], have been widely adopt in scene text detection. According to the principle of text detection, scene text detection methods can be broadly divided into four categories: traditional-based methods, regression-based methods, segmentation-based methods and graph convolution network-based methods.

**Traditional-based methods** Methods based on sliding windows mainly build and move multi-scale windows on the images, and classify the current window into text or non-text. Connected component (CC) based methods firstly get characters candidates by low-level features (such as gradient or color), and then the candidates are classified into text or non-text via the classification model, which include Stroke Width Transform (SWT) [2] and Maximally Stable Extremal Regions (MSER) [6]. However, these traditional-based methods usually adopt a bottom-up strategy and split text detection into multiple steps, including character detection, text line construction and classification. Due to the cumbersome process, these methods perform badly in text detection.

**Regression-based methods** The regression-based methods regard the text instance as an object, and directly estimate the bounding boxes as the results. One-stage detectors of regression-based methods include TextBoxes++ [7], SegLink [15], TextBoxes [8], EAST [29]. Two-stage detectors of regression-based methods include CTPN [17], FEN [27], R2CNN [5] and RRPN [14]. One-stage approaches have gained much attention over two-stage methods due to their competitive performance and simpler design. However, these methods have structural limitations for detecting irregular scene text.

**Segmentation-based methods** The segmentation-based methods detect texts at the pixel level. [10] proposed a Character Attention Fully Convolutional Network (CA-FCN), which adopted an attention mechanism for characters to predict the position of each character in the image. [19] used the progressive scale expansion algorithm to distinguish adjacent text instances. [1] proposed a novel framework called CRAFT, which was adapted from U-Net to detect text region by exploring each character and affinity between characters. [20] adopted a pixel aggregation method (PA) and aggregated text pixels by similarity vectors. [25] proposed a robust arbitrary shape text detection method, namely TextfuseNet, which extracted and fused three levels of features, including character-, word- and global-level to achieve robust arbitrary text detection. [9] developed the DB module to adaptively set the thresholds for binarization, the module not only simplified the post-processing but also enhanced the performance of text detection. They empirically set the K value (50) as a hyper-parameter.

**Graph Convolution Network-based methods** [28] proposed a unified relational reasoning network for text detection. A local graph established the relationship between the text proposal model via Convolutional Neural Network (CNN) and the deep relational reasoning network via Graph Convolutional Network (GCN), as the model can be trained end-to-end.

## 3 Methodology

### 3.1 Architecture

The network architecture of our method is schematically illustrated in Figure 2. We use the ResNet with a feature pyramid network (FPN) as backbone to extract multi-scale feature maps. The features from backbone network are fed into Global module and RCNN module. We add the k submodule and shift submodule into Global module to further optimize the adaptive performance of the binarization module. All submodules in Global module share the FPN convolutional features, which are concatenated together by element-wise sum component. The RCNN module contains region proposal network (RPN) for generating text proposals, double-path fusion component for fusing two-level features and RoI Head for classification and bounding box regression. In the training period, the k map and the shift map are unsupervised, the probability map and the approximate binary map share the same supervision.

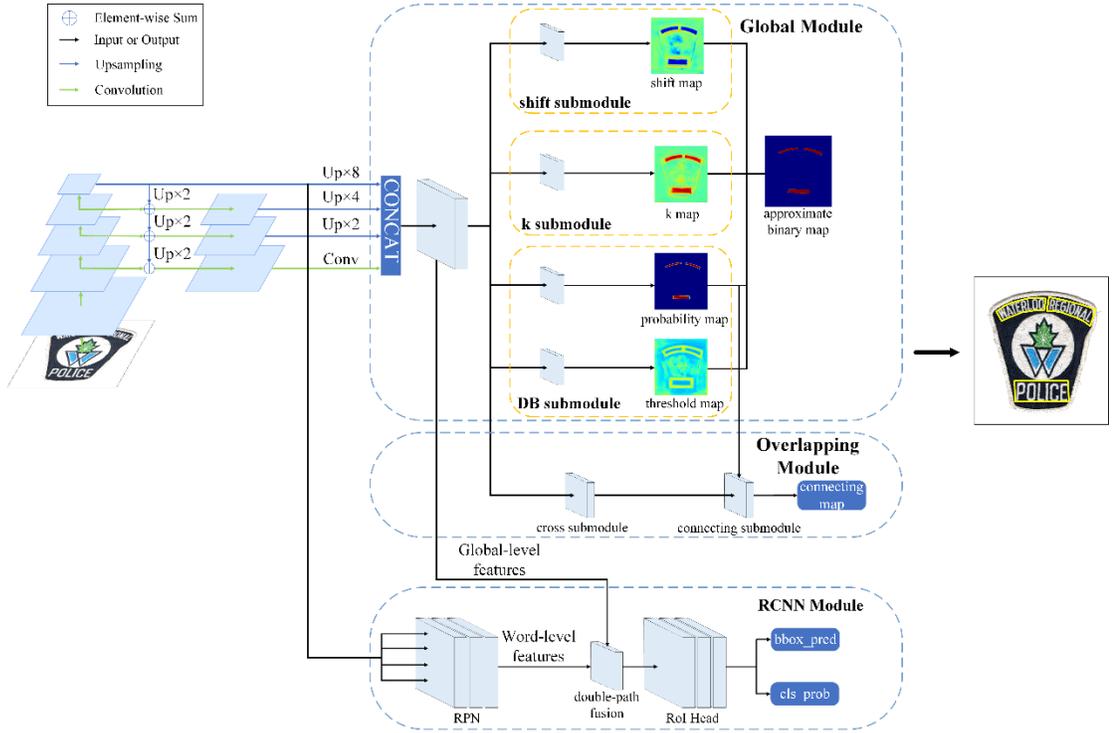

**Fig. 2.** The overall architecture of GWNet. We extract features from global- and word- level for text detection during training period.

### 3.2  GWNet Global module

The Global module is mainly composed by three submodules: DB, k and shift submodule. In the DB submodule, the generating procedures of the probability map ($P$) and the threshold map ($T$) are consistent with DBNet. The k submodule and the shift submodule generate the k map ($K$) and the shift map ($S$), respectively. Then, the approximate binary map ($\hat{B}$) is calculated by $P$, $T$, $K$ and $S$ for exploiting global semantics, and the equation is shown as follows:

$$\hat{B}_{i,j} = \frac{1}{1 + e^{-K_{i,j}(P_{i,j} - T_{i,j} + S_{i,j})}} \tag{1}$$

where $K_{i,j}$, $P_{i,j}$, $T_{i,j}$ and $S_{i,j}$ indicate the $K$ value, the probability value, the threshold value and the shift value of the coordinate point $(i, j)$ in the map respectively.

**GWNet k submodule**  $K$ in formula (1) indicates the amplifying factor, which are used to adjust the gradient, and is set to 50 empirically in DBNet. In this work, we propose the k submodule to enhance the adaptability of $K$ and extract the k map features from FPN to generate the $K$ value. The k maps with different epochs are visualized in Figure 3. With the increase of the training iterations, the *K* values of text regions are close to 70 and the ones of non-text regions are close to 50. Furthermore, we design comparative experiments to show the effectiveness of k submodule. Figure 4 shows the probability map of two methods at different epoch during training periods. As training is performed, the *P* values ($P_{i,j}$) of text regions are gradually increased. More specifically, the P values, which are obtained by "DB+k" submodule, are higher than DB submodule for the middle letters ("*PLAZ*") in natural image. This indicates that the self-adaptive k submodule can achieve a faster convergence speed, and help to avoid falling into local optimal solution. The details of the comparative result are described in the Ablation Study.

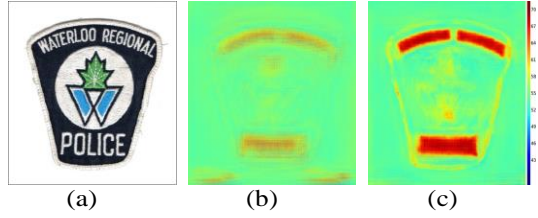

**Fig. 3.** Visualization of the k map with different epochs. (a) The original image. (b) The k map of the previous epochs. (c) The k map of the final epochs.

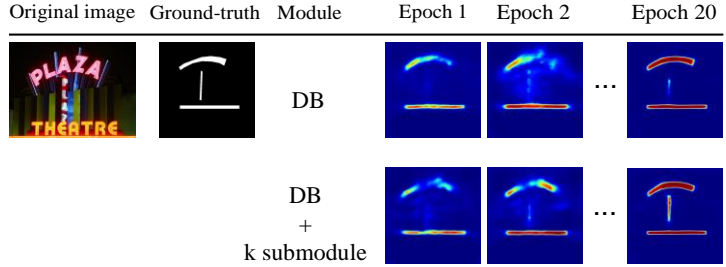

**Fig. 4.** Probability maps with each method during training. Original image means the original training image; Ground-truth indicates the shrunk labels.

**GWNet shift submodule** The input of formula (1) is $P_{i,j} - T_{i,j}$. Formula (1) is sensitive to the change of the $P_{i,j} - T_{i,j}$ value nearby the orign, which means "missing detection" and "false detection" are easy to appear when the values of $P_{i,j}$ are close to $T_{i,j}$. The proposed shift submodule offsets the *P* values. Furthermore, it is mainly to modify the scores of text and non-text regions. The shift submodule can be described as follows:

$$P'_{i,j} = P_{i,j} + S_{i,j}, \quad -0.15 <= S_{i,j} <= 0.15 \qquad (2)$$

where $P'$ is the probability map modified by the shift submodule, $S$ is the shift map learned from the shift submodule.

The shift maps with different epochs are visualized in Figure 5. As shown in Figure 5, with the increase of training iterations, the shift values of text regions and non-text regions will close to -0.1 and 0 respectively. It indicates that shift submodule focuses on negative offsetting the text region scores, which will promote the text region scores enhancement under the constraint of the loss function during training period.

A comparison experiment is set up to show the performance of shift submodule. Figure 6 is a visualization of experiment result on partial testing datasets. The method added shift submodule can predict text regions more accurately, especially for some difficult text regions, such as squiggles or texts with deviation angle, and $P$ values are higher than DB submodule method.

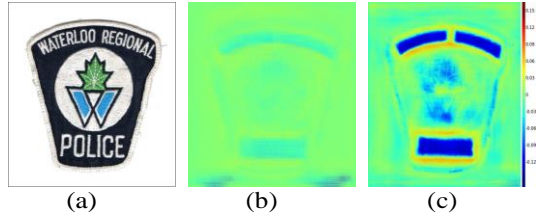

(a)            (b)            (c)

**Fig. 5.** Visualization of Shift map with different epochs. (a) The original image. (b) The shift map of the previous epochs. (c) The shift map of the final epochs.

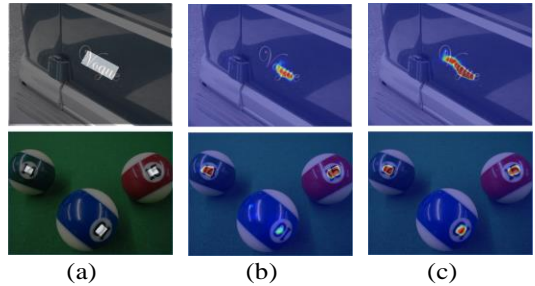

(a)            (b)            (c)

**Fig. 6.** The probability map with/without shift submodule. (a) The shrunk labels. (b) The probability map generated by DBNet. (c) The probability map generated by DB+shift submodule.

### 3.3  GWNet RCNN module

Most existing methods usually seek text regions at a single level, and the single visual information will lead to produce inaccurate detection results. In order to obtain multi-level features for text detection, we propose the RCNN module to extract word-level features and fuse it with global-level features obtained from Global module. Firstly, the fuse features generated by FPN are fed into the RPN submodule to generate text proposals. Then, we introduce a new branch from the Global module to obtain the global-level features. After that, we adopt double-path fusion in global-, word-level features. Figure 7 shows the details of double-path fusion. RoIAlign is applied to extract these two level features. Finally, the double-path fusion features are used for classification and bounding box regression via the RoIHead submodule. The purpose of introducing the RCNN module is to optimize the feature extraction of backbone during training period, as the Global module and the RCNN module share the same backbone, and two modules build connections through double-path fusion.

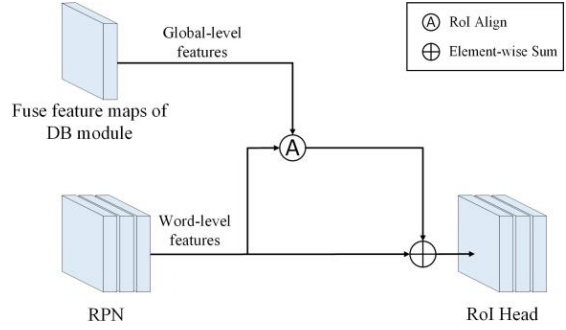

**Fig. 7.** Illustration of the double-path fusion architecture. We fuse global- and word-level features, then feed fuse features into RoI Head.

### 3.4  Ground Truth Label Generation

The label generation includes the global-level and the word-level label generation. Refer to DBNet for more details on the global-level label generation. For the probability map and the binary map, which share the same supervision, the Vatti clipping algorithm is used to generate the ground-truth of text-region by shrinking the polygon $tt$ to $tt_s$. The ground-truth of shrunk text regions can be described as follows:

$$tt_s = \{g_i(x, y)\}_{i=1}^{N} \tag{3}$$

where $g_i(x, y)$ represents the point set of the $i$-th polygon and $N$ is the number of polygons. The ground-truth of word-level is generated from the shrunk text region and is expressed as the minimum axis-aligned rectangle boxes of the text polygons, which is visualized in green rectangles in Figure 8. The ground-truth generation process is shown as:

$$\begin{aligned} tt_b = \{b_i(x_{min}^i, y_{min}^i, w_i, h_i) | \ & 0 < x_{min}^i < W, \\ & 0 < y_{min}^i < H, \\ & t_b < w_i < W, \\ & t_b < h_i < H\}_{i=1}^{N} \end{aligned} \tag{4}$$

$$w_i = x_{max}^i - x_{min}^i, \tag{5}$$

$$h_i = y^i_{max} - y^i_{min}, \tag{6}$$

where $x^i_{min}$ and $y^i_{min}$ indicate the minimum x and y coordinate of $g_i(x, y)$, respectively; $w_i$ and $h_i$ is the width and height of the $i-th$ box; $W$ and $H$ denote the width and height of the image, respectively; $t_b$ is the minimum threshold of the rectangle boundary lengths. The functions like *cv2.boundingRect* provided by OpenCV can be applied for this purpose.

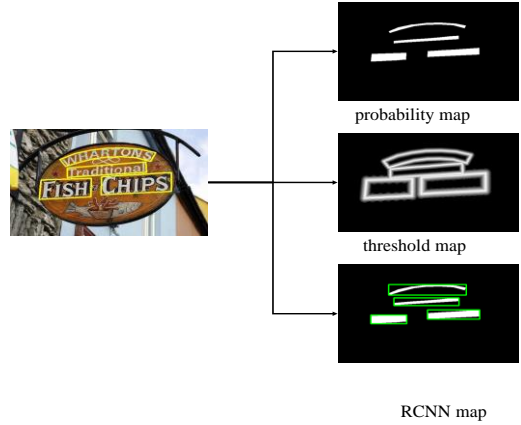

**Fig. 8.** Label generation of the global-level and the word-level label, including probability map, threshold map and RCNN map.

### 3.5  Loss Function

The loss function of GWNet can be expressed as follows:

$$L = \alpha \times (L_p + L_{r\_c} + L_{d\_c}) + \beta \times (L_t + L_{r\_r} + L_{d\_r}) + L_b \tag{7}$$

where $L_p$, $L_t$ and $L_b$ are the loss for the probability map, the threshold map and the binary map, respectively; $L_{r\_c}$ represents the loss for the RPN classification; $L_{r\_r}$ is the loss for the RPN regression loss; $L_{d\_c}$ is the loss for the detection classification; $L_{d\_r}$ is the loss for the detection regression; $\alpha$ and $\beta$ are set to 5.0 and 10.0, respectively.

## 4  Experiments

The experimental datasets include SynthText, Total-Text, CTW1500, MSRA-TD500 and ICDAR 2015.

### 4.1  Implementation details

**Training strategy** The training procedure includes three steps. Firstly, we pre-train the model on SynthText dataset for 80k iterations, then we finetune the pre-trained models on the new datasets. The basic data augmentation for the training data such as flips, rotations and crops are both applied during training period. All the processed images are re-sized to 640×640. The model is optimized using Stochastic gradient descent (SGD) with a weight decay of 0.0001 and a momentum of 0.9. The training batch size is set to 8. The initial learning rate is set to 0.003.

Real-time Scene Text Detection Based on Global Level and Word Level Features    9**Table 1.** Performance contribution of each module in GWNet.

| Backbone | K Module | Shift Module | RCNN Module | MSRA-TD500 P | R | F | FPS | CTW1500 P | R | F | FPS |
|---|---|---|---|---|---|---|---|---|---|---|---|
| ResNet-18* | × | × | × | 90.4 | 76.3 | 82.8(736) | 62 | 84.8 | 77.5 | 81.0(1024) | 55 |
| ResNet-18 | ✓ | × | × | 90.8 | 79.1 | 84.5(736) | 62 | 84.1 | 81.2 | 82.6(1024) | 55 |
| ResNet-18 | × | ✓ | × | 91.2 | 78.7 | 84.5(736) | 62 | 84.2 | **81.3** | 82.7(1024) | 55 |
| ResNet-18 | ✓ | ✓ | × | 91.6 | **79.3** | **85.0(736)** | 62 | **86.0** | 80.6 | **83.2(1024)** | 55 |
| ResNet-50 | ✓ | ✓ | × | 92.0 | 81.7 | 86.5(736) | 31 | 88.6 | 81.8 | 85.1(736) | 24 |
| ResNet-50 | ✓ | ✓ | ✓ | **95.5** | **82.7** | **88.6(736)** | 31 | **89.1** | **85.5** | **87.3(736)** | 24 |

**Table 2.** Performance contribution of each module in GWNet at small scale.

| Backbone | K Module | Shift Module | RCNN Module | MSRA-TD500 P | R | F | FPS | CTW1500 P | R | F | FPS |
|---|---|---|---|---|---|---|---|---|---|---|---|
| ResNet-18* | × | × | × | 90.4 | 76.3 | 82.8(736) | 62 | 84.8 | 77.5 | 81.0(1024) | 55 |
| ResNet-18 | ✓ | × | × | 90.6 | 77.1 | 83.3(640) | **64** | 86.8 | 79.9 | 83.2(640) | **59** |
| ResNet-18 | × | ✓ | × | 89.9 | 76.9 | 82.9(640) | **64** | 86.3 | 78.9 | 82.5(640) | **59** |
| ResNet-18 | ✓ | ✓ | × | **91.9** | **76.9** | **83.7(640)** | **64** | **88.3** | **80.0** | **83.9(640)** | **59** |

**Inference strategy** At the inference stage, the height of the test images are resized to 736 on the MSRA-TD500, 896 on the Total-Text dataset and 704 on the CTW-1500, keeping the aspect ratios. The images are resized to 2048×1152 on the ICDAR2015. We only use the probability map to generate text polygons or text bounding boxes in order to accelerate the inference speed. We inference each dataset using a single 1080Ti GPU, and the batch size is set to 1.

### 4.2 Ablation study

Compared with the DBNet, we introduce two submodules and RCNN module to improve the performance of text detection. Therefore, we conducted an ablation study on MSRA-TD500 and CTW1500 to evaluate how each module in GWNet influences the detection performance. P, R, F and FPS indicate Precision, Recall, F-measure and test speed respectively in Table 1 and Table 2. The values of the bracket indicate the height of the images. "∗" indicates the result of the baseline provided by DBNet.
As is shown in Table 1, the proposed methods achieve more robust and superior performance in detecting arbitrarily shaped texts. For the ResNet-18 backbone, we inference all images with two resolutions. Based on the high inference resolution, the improvements of k submodule and shift submodule are both more than F-measure of 1.7% on MSRA-TD500, while are more than F-measure of 1.6% and 1.7% on CTW1500 respectively. For the ResNet-50 backbone, we make a comparison between the combination of k submodule and shift submodule with/without RCNN module. Compared to the combination without RCNN module, the combination with RCNN module can improves the F-measure by 2.1% and 2.2% beyond the combination without RCNN module on MSRA-TD500 and CTW1500 respectively. The results show that these modules can help to optimize the adaptive performance of the binarization module and extract richer features.
We also make a comparision at small scale, as shown in Table 2. Based on the small-scale detection,the proposed method outperforms the baseline on MSRA-TD500 and CTW1500, in terms of accuracy and inference speed.

**Table 3.** Comparative Results on different datasets.

| Method | Toal-Text | | | | CTW-1500 | | | | ICDAR2015 | | | | MSRA-TD500 | | | |
|---|---|---|---|---|---|---|---|---|---|---|---|---|---|---|---|---|
| | P | R | F | FPS | P | R | F | FPS | P | R | F | FPS | P | R | F | FPS |
| TextSnake [13] | 82.7 | 74.5 | 78.4 | - | 67.9 | 85.3 | 75.6 | 1.1 | 84.9 | 80.4 | 82.6 | 1.1 | 83.2 | 73.9 | 78.3 | 1.1 |
| SPCNet [23] | 83.0 | 82.8 | 82.9 | - | - | - | - | - | 88.7 | 85.8 | 87.2 | - | - | - | - | - |
| CRAFT [1] | 87.6 | 79.9 | 83.6 | - | 86.0 | 81.1 | 83.5 | - | 89.8 | 84.3 | 86.9 | - | 88.2 | 78.2 | 82.9 | 8.6 |
| PSENet [19] | 84.0 | 78.0 | 80.9 | 3.9 | 84.8 | 79.7 | 82.2 | 3.9 | 86.9 | 84.5 | 85.7 | 1.6 | - | - | - | - |
| PAN [20] | 89.3 | 81.0 | 85.0 | 39.6 | 86.4 | 81.2 | 83.7 | 39.8 | - | - | - | - | - | - | - | - |
| DB-ResNet-50* [9] | 87.1 | 82.5 | 84.7 | 32 | 86.9 | 80.2 | 83.4 | 22.0 | 91.8 | 83.2 | 87.3 | 12 | 91.5 | 79.2 | 84.9 | **32** |
| TextFuseNet-ResNet-50 [25] | 83.2 | **87.5** | 85.3 | 7.1 | 85.8 | 85.0 | 85.4 | 7.3 | 91.3 | **88.9** | **90.1** | 8.3 | - | - | - | - |
| CSE [11] | 81.4 | 79.1 | 80.2 | 2.4 | 81.1 | 76.0 | 78.4 | 2.6 | **92.3** | 79.9 | 85.7 | - | - | - | - | - |
| EAST [29] | - | - | - | - | - | - | - | - | 83.6 | 73.5 | 78.2 | **13.2** | 87.3 | 67.4 | 76.1 | - |
| LOMO [26] | 88.6 | 75.7 | 81.6 | - | 89.2 | 69.6 | 78.4 | - | 91.3 | 83.5 | 87.2 | - | - | - | - | - |
| ContourNet [22] | 86.9 | 83.9 | 85.4 | 3.8 | 83.7 | 84.1 | 83.9 | 4.5 | 86.1 | 87.6 | 86.9 | 3.5 | - | - | - | - |
| TextField [24] | 81.2 | 79.9 | 80.6 | - | 83.0 | 79.8 | 81.4 | - | 84.3 | 80.5 | 82.4 | 5.2 | 87.4 | 75.9 | 81.3 | - |
| ATRR [21] | 80.9 | 76.2 | 78.5 | - | 80.1 | 80.2 | 80.1 | - | 89.2 | 86.0 | 87.6 | - | 85.2 | 82.1 | 83.6 | - |
| SAE [18] | - | - | - | - | 82.7 | 77.8 | 80.1 | - | 88.3 | 85.0 | 86.6 | - | 84.2 | 81.7 | 82.9 | - |
| Zhang et al [28] | 86.5 | 84.9 | 85.7 | - | 85.9 | 83.0 | 84.5 | - | 88.5 | 84.7 | 86.6 | - | 88.1 | 82.3 | 85.1 | - |
| **Our(ResNet-50)** | **90.7** | 85.3 | **87.9(896)** | 22 | **89.4** | **85.8** | **87.5(704)** | 26 | 90.6 | 87.8 | 89.2(1152) | 8 | **95.5** | **82.7** | **88.6(736)** | 31 |

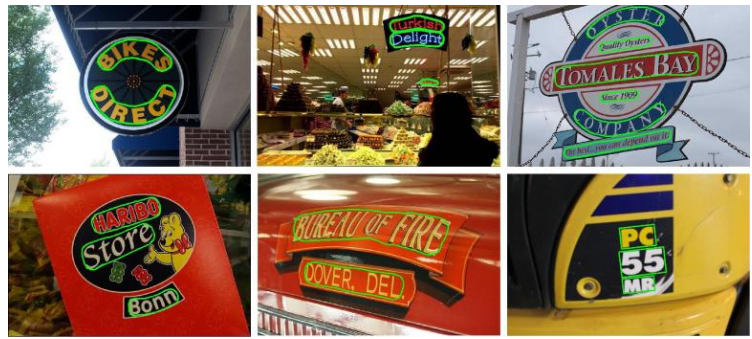

**Fig. 9.** Visualization results on different datasets, including polygon-type and quadrilateral-type.

### 4.3   Comparisons with previous methods

We compared our method with previous methods on four standard benchmarks, including two quadrilateral-type datasets and two polygon-type datasets. Some qualitative results are visualized in Figure 9.

**Polygon-type datasets** CTW1500 and Total-Text are polygon-type text dataset, which contain curved text instances. We can evaluate the performance of the arbitrary shape text detection on these two datasets. The experimental results are shown in Table 3. GWNet enhances the performance and improves the F-measure by 3.2% and 4.1% beyond the baseline on Total-Text and CTW1500 respectively. Moreover, our method achieves state-of-the-art performance on CTW1500 and Total-Text with the backbone of ResNet-50, which outperforms the recent state-of-the-art method by 2.1% and 2.2% on F-measure.

**Quadrilateral-type datasets** We also evaluate the effectiveness of GWNet on quadrilateral-type datasets. All images are inferenced with a single resolution. For MSRA-TD500, we achieve state-of-the-art performance, where the F-measure is 88.6%, which surpass the recent state-of-the-art method [28] by 3.5%, and the precision and recall are 95.5% and 82.7% respectively, which outperforms all pervious methods. As shown in Ablation Study, GWNet with backbone of ResNet-18 can achieve excellent performance without RCNN module during training period. For ICDAR2015, GWNet achieves an F-measure of 89.2% on ICDAR2015, which outperforms the baseline by 1.9%.

## 5  Conclusion

In this paper, we have proposed an arbitrary-shape scene text detector in real-time, which introduces k submodule, shift submodule and RCNN module to optimize the adaptive performance and extract more discriminative representations. The experimental results show that the proposed method has improvement on accuracy compared to previous text detectors, and achieves state-of-the-art performance on three benchmark datasets. In the future, we are interested in detecting cases "text inside text" via RCNN module and developing an end-to-end [16] scene text spotting system.